\documentclass[12pt]{article}
\usepackage[letterpaper, left=1.0in, right=1in, top=1in, bottom=1in]{geometry}
\usepackage{graphicx}  
\usepackage{times}  
\usepackage[bookmarks=true, hidelinks]{hyperref}
\usepackage[page]{appendix}  
\usepackage{rotating}  
\usepackage{enumitem} 
\usepackage{blindtext} 
\usepackage{subcaption}
\usepackage{hyperref}
\usepackage{amsmath}
\usepackage{amssymb}
\usepackage{dsfont}
\usepackage{bbm}
\usepackage{stmaryrd}
\usepackage[labelformat=parens,labelsep=quad, skip=3pt]{caption}
\usepackage{pbox}
\usepackage{changepage}
\usepackage[dvipsnames,usenames]{color}
\usepackage[square,comma,numbers,sort&compress]{natbib}
\usepackage{authblk}
\usepackage{caption}
\usepackage{tikz}
\usetikzlibrary{arrows,automata,arrows.meta}

\usepackage{mdframed}

\usepackage[utf8]{inputenc}%
\usepackage{lmodern}%
\usepackage{textcomp}%
\usepackage{lastpage}%

\mdfdefinestyle{mystyle}{
    backgroundcolor=lightgray!10, 
    innertopmargin=5mm, 
    innerbottommargin=5mm, 
    innerleftmargin=5mm, 
    innerrightmargin=5mm 
}

\usepackage{bold-extra}

\usepackage{amsmath}
\usepackage{amssymb}
\usepackage{amsbsy}
\usepackage{color}

\newcommand{\beq}{\begin{equation}}
\newcommand{\eeq}{\end{equation}}

\newcommand{\beqa}{\begin{eqnarray}}
\newcommand{\eeqa}{\end{eqnarray}}

\newcommand{\beqaN}{\begin{eqnarray*}}
\newcommand{\eeqaN}{\end{eqnarray*}}

\newcommand{\code}[1]{\verb|#1|}

\usepackage{outlines}

\usepackage{hyperref}

\author[1]{Camila Faccini de Lima}
\author[1,2]{Juliano Gianlupi}
\author[1]{John Metzcar}
\author[1]{Juliette Zerick}
\affil[1]{Luddy School of Informatics, Computing, and Engineering, Indiana University Bloomington \-- all authors have contributed equally to the work}
\affil[2]{Corresponding author: jferrari@iu.edu}
\date{\today}

\title{Accelerated solving of coupled, non-linear ODEs through LSTM-AI}

\begin{document}

\maketitle

\begin{abstract}
    The present project aims to use machine learning, specifically neural networks (NN), to learn the trajectories of a set of coupled ordinary differential equations (ODEs) and decrease compute times for obtaining ODE solutions by using this surragate model. As an example system of proven biological significance, we use an ODE model of a gene regulatory circuit of cyanobacteria related to photosynthesis \cite{original_biology_Kehoe, Sundus_math_model}. Using data generated by a numeric solution to the exemplar system, we train several long-short-term memory neural networks. We stopping training when the networks achieve an accuracy of of 3\% on testing data resulting in networks able to predict values in the ODE time series ranging from 0.25 minutes to 6.25 minutes beyond input values. We observed computational speed ups ranging from 9.75 to 197 times when comparing prediction compute time with compute time for obtaining the numeric solution. Given the success of this proof of concept, we plan on continuing this project in the future and will attempt to realize the same computational speed-ups in the context of an agent-based modeling platfom.
\end{abstract}

\section{Introduction}

\subsection{Long-short-term memory neural networks}

When simulating complex biological systems and models, solving differential equations for a long series of time steps can become  computationally expensive.  This may limit parameter space exploration as well as limit the embedding of those models in agent-based models, as is often required in the study of multi-scale and multicellular systems \cite{Metzcar}. Different approaches have been proposed for addressing this issue, such as the work by Parisi \textit{et al.} on solving differential equations with unsupervised NNs \cite{parisi2003}, as well as Rudd and Ferrari on the Constrained Integration Method for solving initial boundary values of partial differential equations \cite{rudd2015}. Here, we follow a similar approach to the one proposed by Wang \textit{et al.} in \cite{Wang2019} by using a long-short-term memory (LSTM)  network.


LSTM networks are a type of Recurrent Neural Network (RNN) proposed by Hochreiter and Schmidhuber in 1997, and work by ``truncating the gradient where this does not do harm" \cite{hochreiter1997}. In this fashion ``LSTM can learn to bridge minimal time lags in excess of 1000 discrete-time steps by enforcing constant error flow through constant error carousels within special units," and ``multiplicative gate units learn to open and close access to the constant error flow" \cite{hochreiter1997}. This type of NN is particularly suited to handle continuous series of data \cite{Wang2019}. Graves and Schmidhuber (2005) explore in-depth applications of such networks and demonstrate its potential for applications in sequence-prediction \cite{graves2005}, making LSTM the natural NN candidate in this work.  

\subsection{Example ODE system}

The ODEs we are using our proof of concept for this method represent a gene regulatory circuit of cyanobacteria related to photosynthesis \cite{original_biology_Kehoe, Sundus_math_model}. They are as follows:

\begin{equation}
    \frac{dA}{dt}  = \frac{V_m}{\kappa_A + 1} - \gamma_A\,\, A
    \label{a_dt}\,\,,
\end{equation}

\begin{equation}
    \frac{dB}{dt}  = \frac{V_B}{\kappa_B + 1} - \gamma_B\,\, B
    \label{b_dt}\,\,,
\end{equation}

\begin{equation}
    \begin{split}
        \frac{dC_{RNA}}{dt} &= \tau_{prc}\,\,\frac{\kappa_A \,\,A}{1 + \kappa_A \,\, A + \kappa_B B}- \gamma_{C_{RNA}} \,\,C_p,
    \end{split}
    \label{c_rna_dt}
\end{equation}

\begin{equation}
    \frac{dC_p}{dt} = \tau_{lrc}\,\, \kappa_A \,\,C_{RNA} - \gamma_{C_P} \,\,C_p\,\,,
    \label{c_p_dt}
\end{equation}

\begin{equation}
    \frac{dZ_{RNA}}{dt} = \frac{V_m\,\,C_p}{\kappa_A + C_p}  - \gamma_{Z_{RNA}} \,\, Z_{RNA}\,\,,
    \label{r_rna_dt}
\end{equation}

\begin{equation}
    \frac{dZ_p}{dt} = \tau_{lrz}\,\, Z_{RNA} - \gamma_{Z_{P}} \,\, Z_{p}\,\,.
    \label{r_p_dt}
\end{equation}

Here, the $\gamma$s are the decay constants for each chemical species, $\tau$s are conversion rates, $V$s are activation factors from blue ($V_m$) light and green ($V_B$) light, and $\kappa$s are conversion rate controllers. $A$ is a gene activated by blue light that initiates protein generation. $B$ is another gene, activated by green light. It functions to deactivate the circuit. $C$ is responsible for ultimately creating the proteins needed for photosynthesis with $Z$ representing those gene products. We use the forward Euler method to solve the equations. We are using $\delta t = 0.01$ minutes as the time-step, and we simulate $50000$ time-steps (i.e. 50 minutes). These solutions form our ground truth for model training and testing. Solving the system using the forward Euler method serves as our computational efficiency benchmark. 

\section{Methods}

\subsection{Data}
Raw data is generated by computationally solving the system of equations. New data sets can be generated at will, and speedily. In approximately 80 minutes using locally available computational power we can generate $10^5$ data sets. To generate the data we uniform-randomly generate each of the parameters ($\gamma$s, $\kappa$s, $V$s) in a range from zero to the maximum typical magnitude in the literature. We also uniform-randomly generate the initial value for each of the equations in a range from zero to the maximum typical magnitude: non-dimensionalized unity (1). We save the entire time series of this system of equations as a matrix, with the first column simulated time and the remainder the current solution. Each row then contains a set of species concentrations at a point in time, which is easily vectorized and further sampled; the first row of course contains the initial conditions.

\subsection{Neural Network}

We deploy a long-short-term memory recurrent neural network (LSTM RNN) to fit the family of coupled ODEs similar to \cite{Wang2019}, noting the previous work applied an LSTM RNN to partial differential equations (PDEs). The exact design of the LSTM RNN and accompanying loss function are described below. We train the NNs with paired vectors from the beginning and end of positive integer multiples of 25 time step-long blocks obtained from among our generated time-series data. During initial experiments we used 80\% of the data set for training, with 10\% for validation and another 10\% for testing. However, in order to satisfy hardware constraints while building a still more diverse and broadly applicable model, both training and testing datasets are uniform-randomly selected with replacement. Due to data set size, the odds of sampling the same vector-pairing for training and testing at each epoch are miniscule.

We use Adam optimization and gradient clipping during training, with summed mean squared error (MSE) as our loss function for the purpose of backpropagation. The evaluation of the goodness of fit of our model is determined by computation of the relative normed error of the testing data set and its model-produced attempt at prediction. This follows the  design seen in \cite{Wang2019}, which we generally followed as a first attempt to build and train our network.

 As mentioned earlier, hardware limitations imposed some constraints which necessitated significant downsampling and purposeful abatement of the speed of training the LSTM model. For construction of datasets for training the LSTM model, input consists of raw data in $10^5$ runs of simulation of a biological switch, with the ability to generate infinitely more. From this, 1000 runs are randomly selected, and for each run, data is sampled every 25 points, or every 0.25 minutes in simulation time, and paired with the Nth sample ahead of it, where N is the “lookahead length”; for this project the goal was to predict with a lookahead length of 25. The first datum is the input vector, and its twin the target (the vector to predict N by 25 time steps later). If a very large dataset is to be constructed, a random selection of these pairs may be performed, but is not advisable to discard data as it makes training with a "shallow" dataset nearly impossible. The samples from all selected runs are pooled, then randomly sampled to create a smaller but still sufficiently varied dataset.

For training the LSTM model, the downsampled dataset is loaded and transformed into a Python 3 iterator. Using Pytorch, an LSTM class object is created with:
\begin{itemize}
\item batch size: 30 (one hot vector encoding)
\item input dimension: 6
\item Hidden dimension: 50
\item Number of layers: 1
\item Output dimension: 6 (the output dimension of a fully-connected linear neural network to map the LSTM result to the target output)
\item Learning rate: 0.0001
\item Loss function: summed MSE 
\item Optimizer: Adam
\end{itemize}

For each epoch, the dataset iterator randomly selects from the dataset, then constructs training and testing tensors. Naturally, convergence of the model is very slow and the model continues to train until relative normed error is less than 3$\%$. This value balances training time with accuracy in the context of this proof of concept model. Greater accuracy can be achieved as needed but requires additional training time.

\begin{figure}
    \centering
    \includegraphics[width=0.9\textwidth]{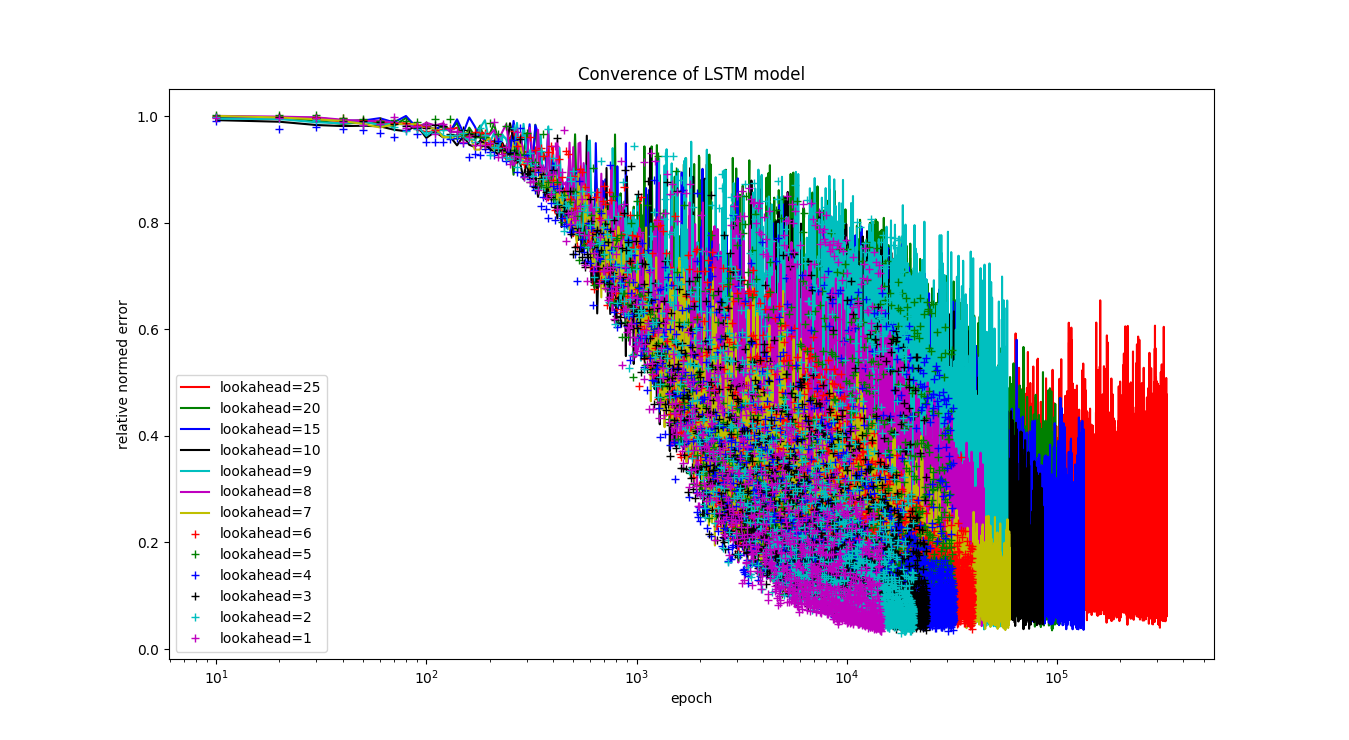}
    \caption{Convergence of LSTM Model}
    \label{fig:training_curves}
\end{figure}

\section{Results}

Figure \ref{fig:training_curves} shows the training curves (given as relative norm error) for multiple values of lookahead length N. We observed that models trained on data created with small values of N converged to satisfactory NN-model parameters (parameters required to reach the set error tolerance) in significantly fewer epochs in comparison to the other attempts. In other words, as the lookahead length increases, so does the number of epochs required to bring down the relative error to acceptable values.

\begin{figure}[htb]
    \centering
    \includegraphics[width=0.9\textwidth]{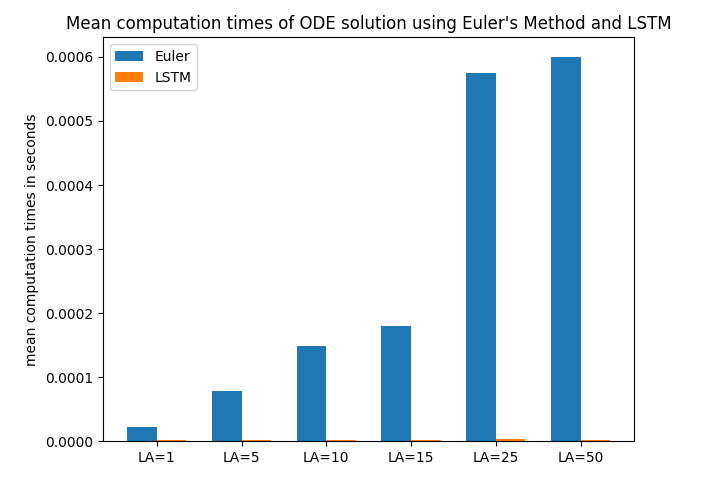}
    \caption{Compute time comparison between neural network prediction and numeric solution via forward Euler.}
    \label{fig:computation_comparison}
\end{figure}

Additionally, we compared the wall time of predicting the a value in the time series using the trained NNs with the compute time of stepping the system of equations through the forward Euler method. As shown in Figure \ref{fig:computation_comparison}, prediction of the next value in the time series (at the various lookahead intervals) via the trained NNs was significantly faster than obtaining the numeric approximation to the original ODEs. Each value is the mean of 10 samplings of the compute time distributions run on a Microsoft Surface Pro, 2017 release. The standard deviations in run times were insignificant. The measured speed ups comparing NN prediction to numeric solving ranged from 9.75 times in the case of a lookahead value of 1 (0.25 min) to 197 times for a lookahead value of 25 (6.25 min). See Table \ref{tab:euler_NN_comparison} for additional details.

\begin{table}[htb]
\centering
\begin{tabular}{ccc}
\hline
 Lookahead value & Euler mean & LSTM mean \\ \hline
 1 &  1.625e-05 &  1.66e-06 \\
 5 & 1.12e-04 & 1.90e-06  \\
10 & 1.14e-04  & 1.43e-06  \\
15 & 3.05e-04  & 1.91e-06  \\
25 & 4.22e-04  &  2.14e-06 \\
50 & 4.99e-04  & 1.91e-06 \\ \hline
\end{tabular}
\caption{Mean value of 10 runs of solving the exemplar set of ODEs in a Jupyter environment contrasted with the mean compute time of predicting the same values using trained neural netorks.}
\label{tab:euler_NN_comparison}
\end{table}

\section{Discussion}


We were able to accomplish our initial goal of training an LSTM to predict values from a family of a set of coupled, non-linear ODEs of biological significance. Additionally, we showed significant computational gains by predicting solutions in the time series of the ODEs via the trained NNs versus stepping the numeric approximations to the ODEs. Given this successful proof of concept, we plan to continue with this method and will next prototype it in the agent-based tissue simulation framework PhysiCell \cite{PhysiCell}, anticipating that we may be able to transfer the efficiency seen in this study to a true multiscale model. 

\section*{Acknowledgments}

We would like to give thanks to Dr. Ariful Azad, Aneequa Sundus, and Dr. Paul Macklin for their help and insights. 


\bibliographystyle{abbrvnat}
\bibliography{main}








\end{document}